\documentclass{article}

\usepackage[final]{neurips_2023}

\usepackage[utf8]{inputenc} 
\usepackage[T1]{fontenc}    
\usepackage{hyperref}       
\usepackage{url}            
\usepackage{booktabs}       
\usepackage{amsfonts}       
\usepackage{nicefrac}       
\usepackage{microtype}      
\usepackage{xcolor}         

\usepackage{amssymb}
\usepackage{mathtools}
\usepackage{amsthm}

\usepackage{graphicx}
\usepackage{wrapfig}
\usepackage{caption}
\usepackage{pifont}
\usepackage{soul,color}
\usepackage{enumitem}
\usepackage{comment}
\usepackage{adjustbox}
\usepackage{bm}
\usepackage{bbm}
\usepackage{makecell}
\usepackage{tcolorbox}
\usepackage{algorithm}
\usepackage{algorithmicx}
\usepackage{algpseudocode}

\definecolor{cgray}{RGB}{234,234,242}

\DeclareMathOperator*{\argmax}{arg\,max}

\newcommand{\cmark}{\ding{51}}%
\newcommand{\xmark}{{\color{black!25}\ding{55}}}

\title{Optimising Human-AI Collaboration by Learning Convincing Explanations}

\author{%
  Alex J. Chan \\
  University of Cambridge\\
  Cambridge, UK \\
  \texttt{ajc340@.cam.ac.uk} \\
  \And 
  Alihan H{\"u}y{\"u}k \\
  University of Cambridge\\
  Cambridge, UK \\
  \texttt{ah2075@cam.ac.uk} \\
  \And
  Mihaela van der Schaar\\
  University of Cambridge\\
  Cambridge, UK \\
  \texttt{mv472@cam.ac.uk} \\
}

\begin{document}

\maketitle

\begin{abstract}
Machine learning models are being increasingly deployed to take, or assist in taking, complicated and high-impact decisions, from quasi-autonomous vehicles to clinical decision support systems. This poses challenges, particularly when models have hard-to-detect failure modes and are able to take actions without oversight. In order to handle this challenge, we propose a method for a collaborative system that remains safe by having a human ultimately making decisions, while giving the model the best opportunity to convince and debate them with interpretable explanations. However, the most helpful explanation varies among individuals and may be inconsistent across stated preferences. To this end we develop an algorithm, \emph{Ardent}, to efficiently learn a ranking through interaction and best assist humans complete a task. By utilising a collaborative approach, we can ensure safety and improve performance while addressing transparency and accountability concerns. Ardent enables efficient and effective decision-making by adapting to individual preferences for explanations, which we validate through extensive simulations alongside a user study involving a challenging image classification task, demonstrating consistent improvement over competing systems.
\end{abstract}

\section{Introduction}

Machine learning (ML) systems and human experts tend to exhibit distinct failure modes when performing a task \citep{fails2003interactive}. In particular, while machine learning systems are often more accurate and efficient than human experts - excelling at detecting subtle patterns that are not obvious to people \citep{fujiyoshi2019deep} - they are prone to failure cases that are hard to detect during training \citep{zhang2019empirical,liu2022practical}, but can lead to obvious test-time mistakes that human experts find trivially easy to correct \citep{yasaka2018deep}. Combine these errors with a high-stakes environment such as criminal justice or healthcare, and the result is an ML system that is dangerous if deployed without oversight. 
The waters are muddied further by a lack of accountability when part (or all) of the decision is made algorithmically,
potentially creating mismatched incentives between developers and end-users \citep{reed2016responsibility}.

\begin{wrapfigure}[32]{R}{0.45\linewidth}
    \centering
    \vspace{-5mm}
    \includegraphics[width=\linewidth]{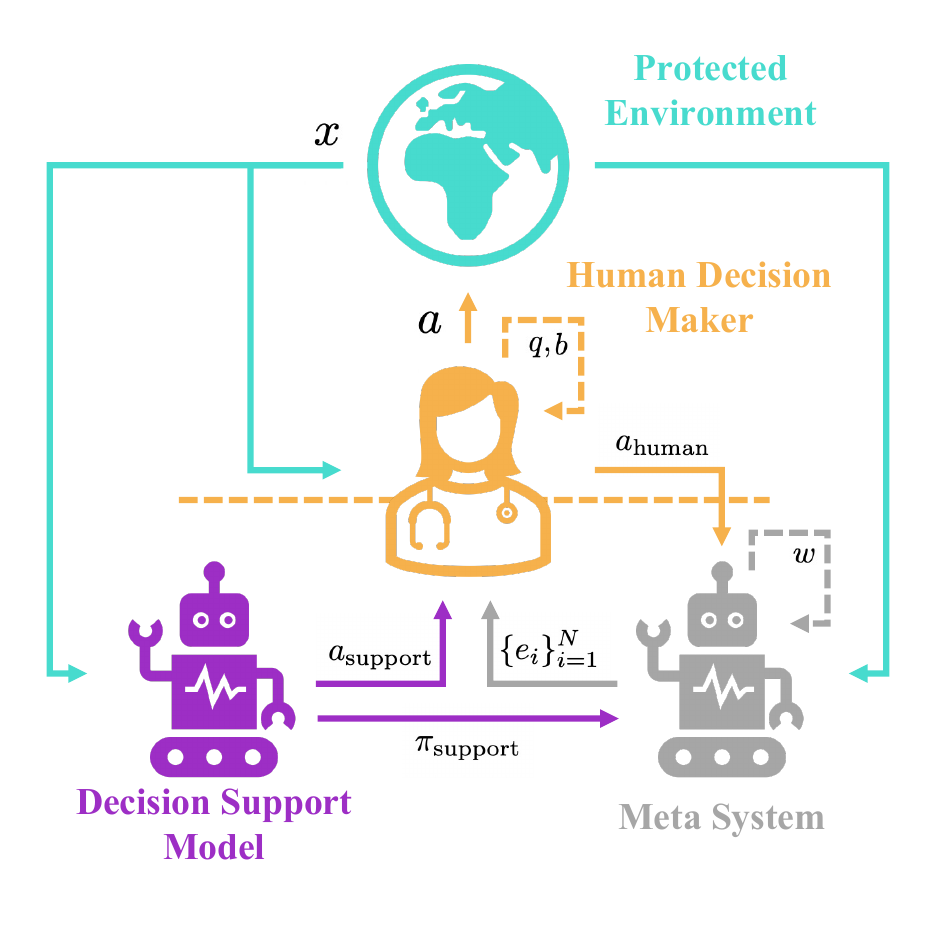}
    \vspace{-7mm}
    \captionof{figure}{\textbf{System Overview.} When interaction with the environment could result in great harm, we would like a system where the human maintains control of actions. We propose \emph{Ardent} as a meta-system built around any decision support model that selects what types of explanations to provide in order to convince the decision maker of its credibility or highlight inaccuracies.}
    \vspace{-1mm}
    \label{fig:overview}

    \centering
\vspace{-1mm}
  \captionof{table}{\textbf{Learning to Select Explanations.} How Ardent compares to other work that explores which explanations to use.} 
  \vspace{0mm}
  \begin{adjustbox}{max width=\linewidth}
  \begin{tabular}{c|cccc}\toprule
    Work & Goal & Feedback & Personalised \\ \midrule
    \cite{lahav2018what} & Interpretability & Trust Score& \cmark \\
    \cite{yeung2020sequential} & Interpretability & Simulatability& \cmark\\
    \cite{wang2021explanations}  & Understanding & Survey& \xmark \\
\midrule
    \textbf{Ardent [US]} & Performance & $a$ only& \cmark \\ \bottomrule
  \end{tabular}
  \end{adjustbox}
  \label{tab:related_exp}
  \vspace{-1mm}
\end{wrapfigure}

A natural solution to this problem is to have a human always be the one to make the decision, while having access to the output of some machine learning model as a decision support tool.
However, even when implemented as support that only \emph{assists} the users, the previous issues can prevent enthusiastic adoption; people often feel like they cannot trust the output of black-box models without any case-specific justification \citep{duran2021afraid}.
Additionally, there is plenty of evidence that the suggestions of the system may psychologically affect the human, shifting their preferences \citep{carroll2022estimating} and potentially manipulating them into taking decisions the system wants - which is unsurprising given it happens to be their stated goal \citep{resnick1997recommender}.

As such, what is needed are systems to guide the interaction between human and machine in order to get the best out of each of them. In this work, 
we propose the development of a decision support system that not only recommends actions, but also actively aims to provide the best possible evidence supporting the credibility of the model's recommendations in order to prevent accurate advice from being dismissed by the human when the rationale behind the advice is not immediately clear.
In order to minimise the chance for manipulation, the type of arguments available to the system are limited to explainability methods \citep{gilpin2018explaining} that offer some insight into the black-box prediction to the human \citep{kenny2021post}, making it easier to identify nonsensical predictions from the model.

We measure the usefulness of explanations based on the eventual agreement of the human with recommended actions, without soliciting explicit feedback from them
as in previous work \citep{wang2021explanations}.
In doing so, we learn if an explanation is truly useful enough to reveal new insight into a model and hence prompt a change in one's behaviour as opposed to merely seeing how interpretable the explanation is \textit{perceived} to be. 
Attempts to learn which explanations should be shown to people are summarised in Table \ref{tab:related_exp}, although in brief have included using Q-learning to learn which explainers to select, but with a reward based on their simulatability score \citep{yeung2020sequential}.
\citet{lahav2018what} on the other hand uses UCB1, an algorithm designed for the standard bandit problem \citep{auer2002finite}, on a reported score from users as to which explanation they \emph{trust} the most. 
The main point of divergence being that these are built around a goal of learning which explanations are \emph{interpretable} - a goal that may not correlate with which are most useful for performance - and as such make use of alternative forms of feedback that may not be appropriate for optimal performance.

\textbf{Paper Roadmap.}
In what follows, we start in Section \ref{sec:system_overview} by developing a framework for meta-system decision support - guiding the interaction between human experts and machine learning support systems.
Based on this framework, we discuss a specific instantiation and potential model of behaviour in Section \ref{sec:model_behaviour} before 
presenting \textbf{Ardent} (\emph{adj. very enthusiastic or passionate}) in Section \ref{sec:ardent}, a method for \underline{Ar}gumentative \underline{de}cisio\underline{n} suppor\underline{t}. Ardent represents a machine learning meta-system, i.e. one that governs the interaction between a human and a decision support system in order to optimise a task.
Finally, in Section \ref{sec:experiments} we demonstrate empirically the benefits of Ardent through a series of experiments.
Here we validate in simulations how useful Ardent can be before putting into practice with real human decision makers in an image classification example.

\section{Meta-systems for Decision Support}\label{sec:system_overview}

In this section we will discuss at a high level the opportunities and challenges faced in building decision support systems for safety-critical tasks.
Consider an arbitrary \textbf{task} $\mathcal{T}$ that needs to be completed by taking some \textbf{action}~$a\in A$ given a \textbf{context}~$x\in X$. We consider the setting where this is some safety-critical task, where ultimately the decision must come down to a \textbf{human} taking actions according to some \textbf{human-policy}~$\pi_{\mathrm{human}}\in\Delta(A)^X$. 
There are two important levels of algorithmic support - we consider a \textbf{decision support system} to be a predictive model with some \textbf{support-policy}~$\pi_{\mathrm{support}}\in\Delta(A)^X$ that is doing the same task as the human, operating on the \emph{same} domain as $\pi_{\mathrm{human}}$. 
On top of this, we consider a \textbf{meta-system} whose task is then essentially to govern the interaction between the two lower-level policies $\pi_{\mathrm{human}}$ and $\pi_{\mathrm{support}}$. This could conceivably take many different forms - for example: occasionally using the human prediction to update the support model; encouraging the human to take the support system more seriously as this context is one that humans often get wrong; or even flagging decisions for an external review.
The overall setup is modelled in Figure \ref{fig:overview}, the key aspect being that it is only ever the human decision maker who is able to \emph{directly} affect the environment. Of course, the support systems are able to influence it \emph{indirectly} (otherwise there would be no point in them), but the human is able to act as a screen to prevent potentially dangerous actions being performed.

\begin{wrapfigure}[19]{R}{0.45\linewidth}
    \centering
    \vspace{-6mm}
    \includegraphics[width=0.9\linewidth]{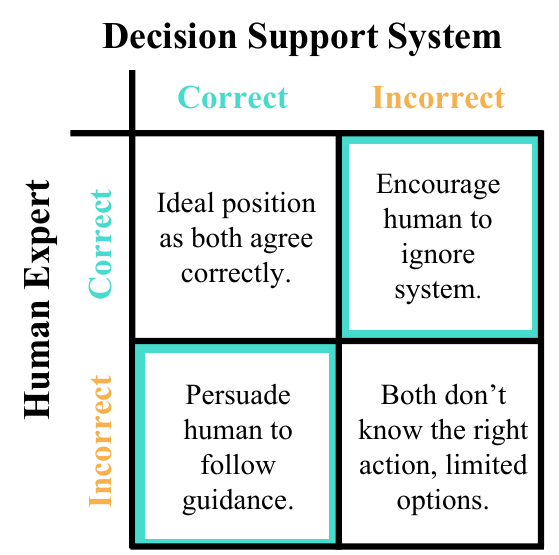}
    \vspace{-1mm}
    \captionof{figure}{\textbf{Situations Faced by a Meta-system:} For each interaction, a meta-system would like to determine which quadrant they find themselves in - a very challenging task.}
    \label{fig:quadrants}
\end{wrapfigure}

\textbf{Identifying Who's Correct.} Similar to problems of learning to defer \citep{mozannar2020consistent} or switch between policies \citep{meresht2020learning},
a key role of the most general meta-policy is essentially to detect who out of the human and support model is making a correct prediction and who is not - resulting in basically four possibilities as highlighted in Figure \ref{fig:quadrants}. We would expect the actions of the system to be heavily dependent on the situation. For example, 
if the system thinks they are in top right, where it thinks the human is correct but the system may not be, it might want to intervene to prevent the human from being swayed by the prediction, for example by highlighting that similar contexts were not common in the support model's training data.
On the other hand, if the system thinks they are in the top left, where both the human and support policy is correct, their job is significantly easier and there is no point wasting time by offering extra justification or caveats. That isn't to say nothing can be done though, as the system could still use the incoming examples for semi-supervision or for improved representation learning.

\textbf{Debate Given Disagreement.}
In essence this gives rise to a debate between the human expert and the support model - albeit one highly skewed towards the human given they are also the judge (the human has no actual \emph{need} to convince the support model). 
This can be seen to have a lot of benefits, with debate allowing for better convergence to optimal actions between agents \citep{ehninger2008decision} and has been proposed itself as a framework for safe artificial intelligence \citep{irving2018ai}.

\textbf{Recommender Systems.} 
A popular category of decision support can be classified as recommender systems. However the typical use of these systems, especially used commercially \citep{shani2005mdp}, relies on convincing the human to pick the option that the model wants \citep{pu2011user}. This essentially assumes that $\pi_{\mathrm{support}}$ strictly dominates $\pi_{\mathrm{human}}$ and thus basically tries to alter $\pi_{\mathrm{human}}$ to converge to $\pi_{\mathrm{support}}$.
In the case where humans are adding value this is highly undesirable, and can have serious effects on the human preferences \citep{carroll2022estimating} as a by-product.
Further, recent work by \citet{vodrahalli2022uncalibrated} has even showed that miscalibration (in particular overconfidence) of a machine learning model's predictions resulted in humans being more likely to accept the suggested actions.
This raises questions about the ethics of \emph{deliberately} inducing overconfidence in a model in a high-stakes environment, making the model mislead the human in an effort to persuade them.

\textbf{Understanding Human Decision Making.} In order to best assist a human decision maker it can be useful to model the decision making behaviour of the individual \citep{jarrett2021inverse}. This can involve using imitation learning or inverse reinforcement learning to model their behaviour \citep{pace2021poetree,chan2021scalable}, or trajectory modelling if we believe their policy is updated over time \citep{huyuk2022inverse,chan2021inverse}. Once a model has been obtained, the support model can be designed to specifically aid the shortcomings of the human policy. These often need simulations to verify though \citep{chan2021medkit}, and having a full model is not always necessary to improve the whole system performance.

\section{A Model of Human Behaviour}\label{sec:model_behaviour}

In this work, we will focus on a design of the meta-policy in a slightly more restricted setting, proposing a method for when there is disagreement between policies (highlighted quadrants in Figure \ref{fig:quadrants}).
We will often expect some disagreement, the support policy
is unlikely to be adopted as the human expert's policy outright, not least because it is most likely a black-box model and hence the human might need to be persuaded of the target policy's credibility.
We refine the setting of Section \ref{sec:system_overview} by considering that there is a set of post-hoc \textbf{explainers}~$E$ at our disposal. Given a context~$x$ and a support-policy~$\pi$, each explainer $e\in E$ can output an explanation $f_e(x,\pi)$. 

Our goal is to develop a \textbf{meta-policy} that simultaneously learns and selects (\emph{cf.}\ explores and exploits) the best explanations to show to the human that are maximally useful to them in order to make their final decision.
Suppose the human is wrong and the support-policy is right, these explainers should be able to sufficiently justify their decision to the human so that they adopt the action. On the other hand, if the support-policy is wrong but the human is right, the explainers should highlight that the support model is making nonsensical predictions, encouraging the human to ignore it.
We consider an interaction loop between the human, support-policy, and meta-policy that goes as follows:
\begin{tcolorbox}[colframe=white,colback=cgray!80]
\begin{enumerate}[leftmargin=5pt,nosep]
    \item A new context $x$ arrives.
    \item The human expresses an intended action $a_{\mathrm{human}}$.
    \item The support policy proposes the same or different action $a_{\mathrm{support}}\sim\pi_{\mathrm{support}}(x)$.
    \item The meta-policy provides a set of explanations $f_e(x,\pi_{\mathrm{support}})$ that are given by explainers $e\in\{e_1,e_2,\ldots\}$ in a specified order, as long as the agent keeps interacting.
    \item The human ends the interaction and takes a final action $a$, which might not necessarily be their intended $a_{\mathrm{human}}$ nor the proposed $a_{\mathrm{support}}$.
\end{enumerate}
\end{tcolorbox}
To be able to make meaningful inferences regarding how the system's explanations have influenced the human's final action, we need to model how the human reasons about the information provided by the explanations. In particular, we need to model (i) how they accumulate information as they see multiple explanations one after another and (ii) how they then decide on a final action.

Given a context~$x$, suppose the human considers there to be an optimal action $a^*(x)$ to take but they are not absolutely certain what that action might be. Their policy (i.e.\ the human policy $\pi_{\mathrm{human}}$) reflects their initial belief regarding the optimal action---that is they believe $a^*(x)=a$ to be the case with a confidence of $\pi_{\mathrm{human}}(x)[a]$. We will denote this initial belief with $b_1\in\Delta(A)$ where $b_1=\pi_{\mathrm{human}}(x)$. The agent updates their belief as they gather more information by interacting with the system. Formally, when they are provided with the $t$-th explanation $f_{e_t}(x,\pi_{\mathrm{support}})$ by the $t$-th explainer~$e_t$, they update their belief such that:
\begin{equation}
    b_{t+1}[a] \propto b_t[a]\cdot t \cdot q[e_t,x,a] ~,
\end{equation}
where $q[e_t,x,a]\in\mathbb{R}_+$ can be interpreted as a measure of how likely the agent thinks they are to see the information provided by explanation $f_{e_t}(x,\pi_{\mathrm{support}})$ if $a^*(x)=a$ were to be true---in other words, $q[e,x,a]\propto\mathbb{P}(f_e(x,\pi_{\mathrm{support}})|a^*(x)=a)$. Finally, when the agent ends the interaction with the system after seeing the $T$-th and the final explanation, they take an action~$a$ according to their final belief~$b_{T+1}$ such that $a\sim b_{T+1}$.

\textbf{Objective.}
Our objective is to find a strategy to select explainers $\{e_1,e_2,\ldots\}$ given a context~$x\in X$ and the agent's intended action~$a_{\mathrm{human}}\in A$ according to data $\mathcal{D}=\{(a_{\mathrm{human}},a,e_{1:T})\}$ collected during previous interactions so that the number of times the proposed action is taken as the final action (i.e.\ $a=a_{\mathrm{support}}$) is maximised. We consider the case when propensities~$q\in\mathbb{R}_+^{E\times X\times A}$ and the human policy~$\pi_{\mathrm{human}}$ are unknown.

\section{Argumentative Decision Support}\label{sec:ardent}

Having established the forward model of behaviour we posited in the previous section,
we now present \textbf{Ardent}, a method for \underline{ar}gumentative \underline{de}cisio\underline{n} suppor\underline{t}.
As an \textit{online} learner, Ardent has to strike a balance between two conflicting objectives: (i) infer how explanations affect the human's beliefs by trying out a variety of explanations (i.e.\ \textit{exploration}), and (ii) help the human by showing them only the best explanations (i.e.\ \textit{exploitation}). To achieve this, we employ a variation of Thompson sampling \citep{russo2018tutorial}, a common method for online learning. For each interaction, Ardent first forms a posterior $\mathbb{P}(q|{x,e_{1:T},a})$ over unknown propensities given information from previous interactions. Then, it selects explanations as if a particular sample $q^*\sim \mathbb{P}(q|{x,e_{1:T},a})$ from the formed posterior is the ground-truth propensities.
%

\textbf{Posterior Inference.} 
Since Ardent is intended to be a lifelong learner, it needs to be able to form posteriors over propensities without having to repeatedly retrain a system. This amounts to performing Bayesian updates every time an interaction occurs given an appropriate starting prior.

Given a prior distribution $\mathbb{P}(q)$ over propensities $q\in\mathbb{R}^{E\times X\times A}$, the posterior distribution after observing an interaction where the context is $x$, explainers~$e_{1:T}$ are shown to the agent, and the agent has taken the final action~$a$ can be expressed as:
\begin{equation}
    \begin{aligned}
        \mathbb{P}&(q|x,e_{1:T},a)
        \propto \mathbb{P}(q)\mathbb{P}(a|x,e_{1:T},q)
        = \mathbb{P}(q)b_T[a] 
        = \mathbb{P}(q)\frac{b_1[a]\prod_{t\in[T]}t \cdot q[e_t,x,a]}{\sum_{a'\in A}\big(b_1[a']\prod_{t\in[T]}t \cdot q[e_t,x,a']\big)} ~.\nonumber
    \end{aligned}
\end{equation}


\begin{wrapfigure}[34]{R}{.5\linewidth}
    \vspace{-3mm}
    \captionof{algorithm}{Ardent}\label{alg:ardent}
    \textbf{Input:} Prior distribution $\mathbb{P}(q)\in\Delta(\mathbb{R}^{E\times X\times A})$, and discount factor $\alpha\in(0,1)$
    \begin{algorithmic}
        \State $\forall i\in[N],\: q^{(i)}\sim\mathbb{P}(q)$
        \State $\forall i\in[N],\: w^{(i)}\gets 1/N$
        \Loop
        \Statex \vspace{3pt}\textit{Interaction:}\vspace{3pt}
            \State Context $x\in X$ arrives
            \State Determine action $a_{\mathrm{target}} \sim \pi_{\mathrm{target}}(x)$
            \State $k \sim \mathcal{C}(w^{(1:N)})$ \hspace{12pt}$\triangleright$ \textit{Posterior sampling}
            \Repeat ~\textbf{for} $t\in\{1,2,\ldots\}$
                \State $e_t \gets \argmax_{e\in E\setminus\{e_1,\ldots,e_{t-1}\}}$ \\
                \hfill$q^{(k)}[e,x,a_{\mathrm{target}}]$
                \State Show explanation $f_{e_t}(x,\pi_{\mathrm{target}})$
            \Until{the final action is taken}
            \State Observe the final action $a\in A$
        \Statex \vspace{3pt}\textit{Posterior update:}\vspace{3pt}
            \State $\bar{q} \gets \sum_{j\in[N]}w^{(j)}q^{(j)}$
            \State $\Sigma \gets \sum_{j\in[N]}w^{(j)}(q^{(j)}-\bar{q})(q^{(j)}-\bar{q})^{\mathsf{T}}$
            \State $\forall i\!\in\![N],\: \mu^{(i)}\!\gets\! \alpha q^{(i)}+(1-\alpha)\bar{q}$
            \State $\forall i\!\in\![N],\: p^{(i)}\!\gets\! w^{(i)}\mathbb{P}(a|x,e_{1:T},q\!=\!\mu^{(i)})$
            \State $\forall i\!\in\![N],\: p^{(i)}\!\gets\! p^{(i)}/\sum_{j\in[N]}p^{(j)}$
            \For{$i\in\{1,\ldots,N\}$}
                \State $k \sim \mathcal{C}(p^{(1:N)})$
                \State $q^{(i)} \sim \mathcal{N}(\mu^{(k)},(1-\alpha^2)\Sigma)$
                \State $w^{(i)} \gets \mathbb{P}(a|x,e_{1:T},q=q^{(i)})$ \\
                \hfill$/\mathbb{P}(a|x,e_{1:T},q=\mu^{(k)})$
            \EndFor
            \State $\forall i\in[N],\: w^{(i)}\gets w^{(i)}/\sum_{j\in[N]}w^{(j)}$
        \EndLoop
    \end{algorithmic}
\end{wrapfigure}

\begin{table*}
  \centering \vspace{0mm}
  \caption{\textbf{Multi-Armed Bandit Related Ideas.} A comparison of how Ardent works placed in the context of multi-armed bandits.} 
  \vspace{0mm}
  \setlength{\tabcolsep}{8pt}
  \begin{adjustbox}{max width=\textwidth}
  \begin{tabular}{c|cccc}\toprule
      Problem & Ref. & Arms & Feedback Type & Feedback Model \\ \midrule
      Standard MAB & \cite{auer2002finite} & Individual & Bandit & \textcolor{black!20}{N/A} \\
      CMAB & \cite{chen2013combinatorial} & Combinatorial & Semi-bandit & Deterministic \\
      Cascading bandits & \cite{kveton2015cascading} & Combinatorial & Semi-bandit & Cascading binary choices\\
      CMAB-PTA & \cite{huyuk2019analysis} & Combinatorial & Semi-bandit & Possibly stochastic \\
      MNL-Bandit & \cite{agrawal2019mnlbandit} & Combinatorial & Full-bandit & Multinomial logit (MNL) choice \\
\midrule
    \textbf{Ardent} & \textbf{[US]} & Combinatorial & Full-bandit & Cascading MNL choices \\ \bottomrule
  \end{tabular}
  \end{adjustbox}
  \label{tab:related_bandit}
  \vspace{-8mm}
\end{table*}

Note that it is not possible to keep an analytical track of this posterior, unlike typical applications of Thompson sampling. This is a direct consequence of our feedback model; our aims is to learn solely from the final action~$a$ without relying on explicit feedback from the human. For instance, if we were able to observe $q[e_t,x,a]$'s directly (perhaps by asking the human to score each explanation numerically or express their beliefs at each step explicitly), we could have assumed $\mathbb{P}(q)$ is Gaussian and trivially obtained $\mathbb{P}(q|x,e_{1:T},a)$.
Rather than keeping an analytical track of the posteriors, we perform approximate posterior sampling using a sequential Monte Carlo method instead. In particular, building on the algorithm proposed by \citet{liu2001combined} which outlines how to track distributions over general static parameters such as $q$. We represent distributions over propensities~$q$ with particles $\{q^{(i)}\}_{i\in[N]}$ and their corresponding weights $\{w^{(i)}\}_{i\in[N]}$ such that $w^{(i)}\geq 0,\forall i\in[N]$ and $\sum_{i\in[N]} w^{(i)}=1$. Algorithm~\ref{alg:ardent} describes in detail how these particles are updated. We denote with $\mathcal{N}(\mu,\Sigma)$ the Gaussian distribution with mean vector $\mu\in\mathbb{R}^d$ and covariance matrix $\Sigma\in\mathbb{R}^{d\times d}$, and with $\mathcal{C}(p)$ the categorical distribution over $\{1,\ldots,d\}$ with event probabilities $p\in[0,1]^d$.

\textbf{Explanation Selection.}
Now at a new time-step Ardent has a constructed posterior over the human's beliefs and given a new context and support system prediction is tasked with selecting appropriate explainers to show to the human.
To do so, a particle is sampled from the posterior distribution according to its relative weight ($q^{(k)}: k\sim\mathcal{C}(w^{(1:N)})$ in Algorithm~\ref{alg:ardent}). Then, the explainers are shown to the human in order of their propensity---that is explainers with the largest $q^{(k)}[e,x,a_{\mathrm{target}}]$ are show first---as long as the human continues to request further explainers.

\textbf{Learning from Logged Feedback.}
While Ardent is primarily designed to run online, if there is logged data available about the interaction of the human expert with the decision support tool previously - say collected when shown random explanations - this information can be easily incorporated in order to build an informative prior for the propensities (for instance, by sampling initial particles via Markov chain Monte Carlo methods) before Ardent is deployed.


\textbf{Relationship to Multi-Armed Bandits.}
Ardent is a potential solution to a \textit{combinatorial multi-armed bandit} problem with full-bandit feedback, unlike those with semi-bandit feedback that have been studied extensively. In our framework, semi-bandit feedback would correspond to observing propensities~$\{q[e_t,x,a]\}_{t\in[T]}$ directly in addition to the final action~$a$. Some work considers a special case of full-bandit feedback where observations are dictated by a multinomial logit (MNL) choice model. When all interactions involve only one explanation (i.e.\ $T=1$), our observation model becomes equivalent to theirs. Therefore, our framework could be considered as a generalisation of theirs at least from a technical point of view, although conceptually the two frameworks aim to solve completely different problems. Ardent can be thought of as a \textit{learning-to-rank} problem as our strategy essentially aims to order explanations based on propensities $\{q[e,x,a_{\mathrm{support}}]\}_{e\in E}$ for a given context~$x$ and a given action~$a_{\mathrm{support}}$. However, learning-to-rank problems are typically formulated as problems with semi-bandit feedback---rather than full-bandit feedback---and do not typically feature the complication of observations being dictated by a logistic model---as in our case. A comparison on how similar systems to Ardent might be implemented using alternative bandit frameworks is given in Table \ref{tab:related_bandit}.

\vspace{-3mm}
\section{Experimental Demonstrations}\label{sec:experiments}
\vspace{-2mm}
Now that we have introduced Ardent as a meta-system for decision support, in this section we will explore practically how it works and can be useful. We start by validating its efficacy on a simulated synthetic scenario, before testing it on a real image classification task.

\vspace{-4mm}
\subsection{Validation with Synthetic Agents}
\vspace{-2mm}

Before we consider experiments involving real people making any decisions we will first validate Ardent in a synthetic setting so as to confirm that it behaves as expected as well as examine the effects of different variables on the performance of the system as a whole.
To begin in the simplest case, we will consider a scenario with binary contexts, binary actions, and a binary selection of explanations available to Ardent. Since we focus on ``high-stakes'' environments, we might 
consider a diagnostic setting, where patients either have some disease or not. There are two populations: Patients with context $x=0$ are usually healthy and do not need a treatment $a^*(x=0)=0$, and patients with context $x=1$ who are susceptible to the disease and consequently will require treatment $a^*(x=1)=1$.
Now, in this case the human expert clinician is able to make accurate decisions for $x=0$ (with high probability), specifically $\pi_{\mathrm{human}}(x=0)[a=0]=0.9$, but is unable to do so for $x=1$; they effectively take random actions, specifically $\pi_{\mathrm{human}}(x=1)[a=1]=0.5$. 
The machine learning system on the other hand, is the opposite; they are accurate for $x=1$ but decide randomly for $x=0$: $\pi_{\mathrm{support}}(x=1)[a=1]=0.9$ and $\pi_{\mathrm{support}}(x=0)[a=0]=0.5$. The clinicians believe in their ability and cannot be persuaded of anything when they are certain of their decision (when $x=0$), and further only one of two potential explanations can persuade them to take action $a=1$ when $x=1$. Formally, $E=\{e_{-},e_{+}\}$ and $q[e_+,x=1,a=1]=10$ but $q[\cdot,\cdot,\cdot]=1$ otherwise.

\textbf{System Performance.}
How do various systems fare at the task? We compare the following:\vspace{-2mm}
\begin{tcolorbox}[colframe=white,colback=cgray!80]
\begin{itemize}[leftmargin=0pt,nosep]
    \item \textbf{Human - Alone}: Only the human expert acting.
    \item \textbf{Machine - Alone}: Only the decision support acting.
    \item \textbf{Human + Machine with Random Explanations}: The human is shown the support prediction with a random explanation and then makes a decision.
    \item \textbf{Human + Machine with Oracle Explanations}: The human is shown the decision support system prediction along with the explanation an ``Oracle'' knows will convince them if appropriate, and then makes a decision.
    \item \textbf{Human + Machine with Ardent}: The human is shown the decision support system prediction along with an explanation chosen by Ardent, and then makes a decision.
\end{itemize}\vspace{-1mm}
\end{tcolorbox}

\begin{wrapfigure}[12]{R}{.5\linewidth}
\vspace{-4mm}
    \captionof{table}{\textbf{Accuracy.} $a\to b$ denotes a change from $a$ to $b$ over time. Human+Machine with Ardent eventually achieves the best possible accuracy for both contexts.}
    \vspace{-1mm}
    \resizebox{\linewidth}{!}{\begin{tabular}{@{}lcc@{}}
        \toprule
        \bf Algorithm & \bf\makecell{Accuracy\\for $\bm{x=0}$} & \bf\makecell{Accuracy\\for $\bm{x=1}$} \\
        \midrule
        Human - Alone & $\bm{90\%}$ & $50\%$ \\
        Machine - Alone & $50\%$ & $90\%$ \\
        H+M w/ Random Explanations & $\bm{90\%}$ & $75\%$ \\
        H+M w/ Oracle Explanations & $\bm{90\%}$ & $\bm{95\%}$ \\
        \midrule
        H+M w/ \textbf{Ardent} & $\bm{90\%}$ & $75\%\to\bm{95\%}$ \\
        \bottomrule
    \end{tabular}}
    \label{tab:synth_results}
    \vspace{-0mm}
\end{wrapfigure}

\begin{figure*}
\centering
\vspace{-3mm}
\includegraphics[width=1.0\textwidth]{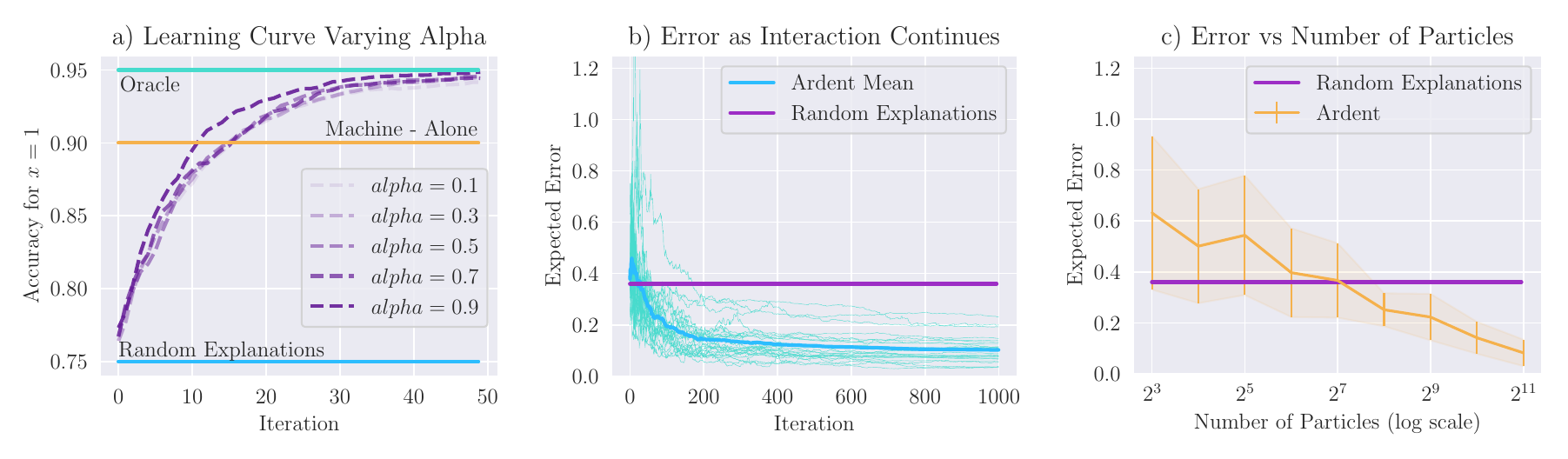}
\vspace{-7mm}\caption{\textbf{Simulated Ablations.} We demonstrate through simulation that \textbf{a)} Ardent is able to rapidly converge on oracle performance. \textbf{b)} As dimensions increase convergence is slower but still very quickly outperforms random explanations. \textbf{c)} Given the approximate nature of inference, the expected error reduces with order of the log number of particles in the filter.}
\label{fig:ablation} 
\vspace{-7mm} 
\end{figure*}

The resultant accuracy for all systems is reported in Table \ref{tab:synth_results}. Ardent starts at, and maintains, an optimal $90\%$ accuracy for $x=0$ as the human is able to always select the action they think is best. For $x=1$, Ardent starts at the same ability as random explanations (and above the human alone), before rapidly overtaking the performance of the isolated decision support model and converging on the oracle performance.
The speed of convergence for Ardent to $95\%$ in the setting where $x=1$ can be seen in Figure \ref{fig:ablation}\textbf{a}. It takes minimal interaction until Ardent is able to select the correct explanation reliably for a wide range in values of $\alpha$.
In conclusion:
\emph{Ardent maintains the benefits of a human in
control while improving overall accuracy after minimal interaction.}

\textbf{Understanding Approximation Impact.}
We consider a generalisation of the previous simulated example with $E=2, X=3, A=4$, where distributions are randomly sampled, with unnormalised logits Normally distributed.
As discussed in Section \ref{sec:ardent}, Ardent employs an approximate Bayesian method in the form of a particle filter, and so considerations have to be made as to how well this can actually track the posterior and allow for accurate performance.
In Figure \ref{fig:ablation}\textbf{b} we can track the accuracy under individual particles as they are updated, as well as the expected value and see that they rapidly outperform the random explanation baseline. 
In Figure \ref{fig:ablation}\textbf{c} we plot how the error reacts to the number of particles in the filter - a key hyperparameter choice when it comes to sequential Monte Carlo methods. We can see that with too few particles the approximation is too coarse and is unable to perform well at the task, although after about 1000 we can be confident in outperforming the baseline. There is of course a trade-off in that the more particles that are simulated, the more that need to be tracked and the higher the computational burden that comes with the increased fidelity.
To summarise: \emph{Expected error reduces rapidly and with order of the log number of particles in the filter.}

\subsection{Challenging Humans with Image Classification}

Having validated Ardent in simulation, we now move on to one example of how this could be used in practice by human decision makes to complete a task, albeit tested in a slightly lower-stakes environment than we describe previously.
CIFAR-10 has been a very common multi-class classification benchmark in the computer vision community \citep{krizhevsky2009learning}, although recently has been largely set aside for bigger and higher resolution image datasets. However, it is the low resolution of CIFAR-10 that makes it a particularly appropriate task for our purposes, as it can still pose a challenge for human labellers, and deep neural networks can achieve very strong accuracy \citep{dosovitskiy2020image}.
The CIFAR-10 test set contains 10,000 images, although many of them are trivially easy for both humans and machine learning systems.
As such, we construct a more curated test set of only 70 images
while over-representing test examples that humans have trouble identifying and deep networks commonly make mistakes on. In this case the overall performance of both humans and machine on this subset is significantly lower than what might be achieved over the full test set. 
This is important for increasing the number of examples for which there is disagreement between human and machine, better representing the type of tasks we expect Ardent to be useful on.
Details of presentation and test-set specifics are given in the supplementary materials. In total, we recruited $32$ participants and received approval from our department's Ethics Committee (IRB equivalent), following standard data collection protocols. Risk was deemed to be low given the task nature and non-identifiable information collected.
Participants were volunteers sought from our institution.

\begin{figure*}
\centering
\includegraphics[width=1.0\textwidth]{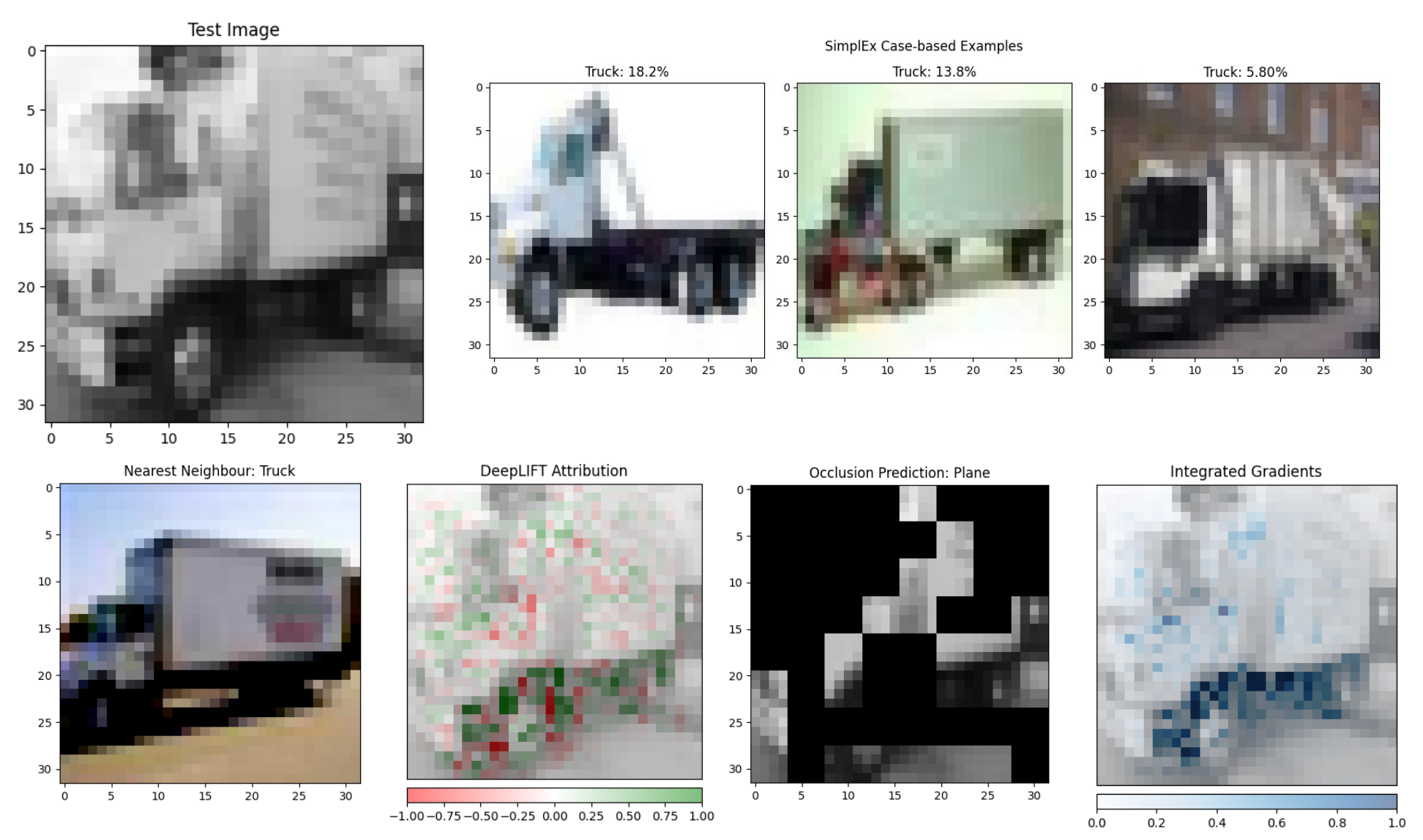}
\vspace{-5mm}
\caption{\textbf{Example Image and Explanations.} Subjects are shown a new test image as in the top left, and asked to make a prediction.
The system then shows them the model prediction, in this case `Truck', and as long as the subject remains unconvinced, continues to show them new explanations - examples of which are shown here.
Details of exact presentation is given in the Appendix. }
\label{fig:example_human} 
\vspace{-7mm} 
\end{figure*}

In order to test the ability of Ardent to optimise performance and discover which explainability methods are preferred by different people we use five different explainability methods that fall in three different categories. This allows for reasonable heterogeneity between explanations, not having them all basically report the same thing.
To that end, we employ:
\textbf{1)} \emph{Feature Importance Methods}: Those that aim to highlight which part of the context was useful for the model in making a decision. In particular we use \textbf{Integrated Gradients} \citep{sundararajan2017axiomatic} - A method for attributing features to a model's predictions while satisfying definitions of sensitivity and implementation invariance; and
\textbf{DeepLIFT} \citep{shrikumar2017learning} - Deep Learning Important Features aims to decompose the prediction into attributions of individual neurons and comparing to a reference attribution to determine feature relevance.
\textbf{2)} \emph{Example Based Methods}: Those that aim to justify the model's prediction by showing other example(s) from a corpus (often the training set) that are in someway similar to the test example including \textbf{SimplEx} \citep{crabbe2021explaining} - that provides relevant examples by reconstructing a test example's latent representation as a mixture of the corpus representations; and
\textbf{Nearest-Neighbour} \citep{wallace2018interpreting} - that provides the example and model prediction of the corpus member closest to the test example in the model latent space.
\textbf{3)} \emph{Counterfactual Methods}: Those that ask a question of the model as to what might the predicition be \emph{if the context had of been different}, in this case \textbf{Occlusion Maps} \citep{zhang1997visibility} that searches for the minimal mask that will result in a different prediction being outputted by the model.
An example of the test images and accompanying examples shown to human experts is shown in Figure \ref{fig:example_human} - note this is not how they are presented during the task, where one explanation would be shown at a time - the actual display shown to participants is detailed in the appendix.
As one can see, all of the different methods offer different information about the decision support model's prediction and so can be useful to different people in different ways, it is very much a \emph{subjective} position as to which one may be more useful.

\begin{wrapfigure}[41]{HR}{.5\linewidth}
    \centering
    \vspace{-4mm}
    \captionof{table}{\textbf{Accuracy.} We report the mean accuracy ($95\%$ confidence interval) on a challenging subset of the CIFAR-10 image classification test set.}
    \vspace{-1mm}
    \begin{adjustbox}{max width=\linewidth}
    \begin{tabular}{@{}lc@{}}
        \toprule
        \bf Algorithm & \bf Accuracy \\
        \midrule
        Human - Alone & $72.5\pm6.2 \%$ \\
        Machine - Alone & $50.0\pm0.0 \%$  \\
        H+M w/ Random Explanations & $76.1\pm3.8 \%$ \\
        H+M w/ \emph{a priori} Favourite  & $75.7\pm4.0 \%$ \\
        \midrule
        H+M w/ \textbf{Ardent} & $\mathbf{83.4\pm5.0 \%}$  \\
        \bottomrule
    \end{tabular}
    \end{adjustbox}
    \label{tab:image_results}
    \vspace{-0mm}
    
    \centering
    \vspace{1mm}
    \includegraphics[width=0.95\linewidth]{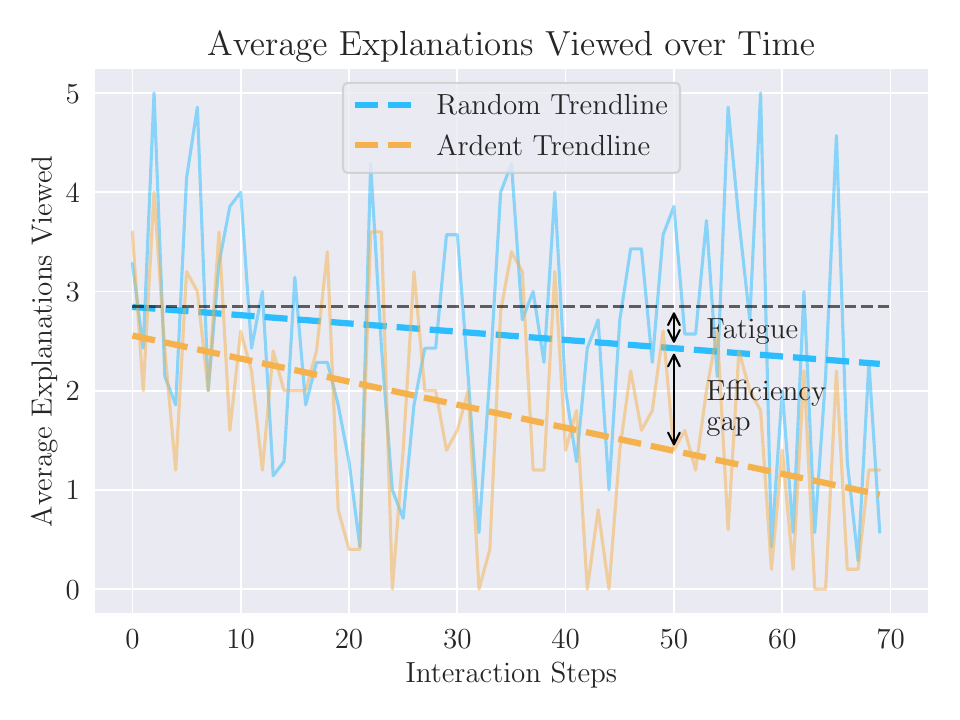}
    \vspace{-1mm}
    \captionof{figure}{\textbf{Explanations Viewed.} Average explanations viewed by participants in the Ardent group and the Random group.}
    \label{fig:viewed}

    \centering
    \vspace{-1mm}
    \includegraphics[width=\linewidth]{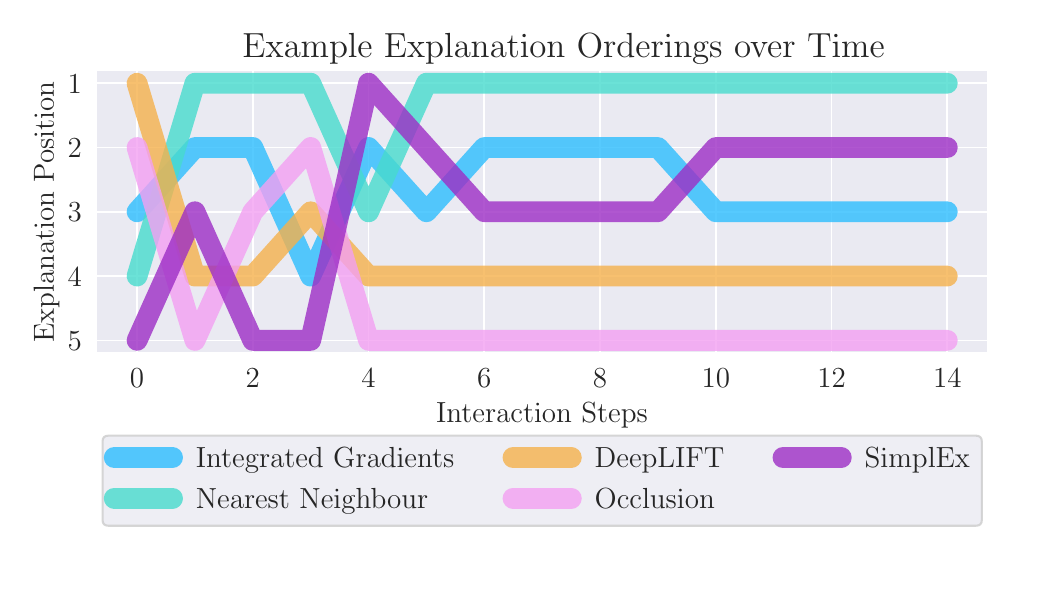}
    \vspace{-8mm}
    \captionof{figure}{\textbf{Preference Inference.} We can see that Ardent quickly identifies that this participant found that example-based explanations were most useful for them.}
    \label{fig:preferences}
\end{wrapfigure}

\textbf{Ability to Accurately Classify Images.} 
All participants were randomly allocated to one of three arms in the trial. These included: \textbf{1)} being shown explanations chosen by Ardent; \textbf{2)} being shown randomly ordered explanations; and \textbf{3)} being shown only the explanation that the participant selected as their favourite at the beginning of the experiment when shown the an example of how the explanations work.
The results for final accuracy on the test set are reported in Table \ref{tab:image_results}, where the estimate of the \emph{Human - Alone} accuracy is calculated from the initial prediction of participants across all arms.
We can see that Ardent significantly outperforms both of the individual (human or AI) systems as well as beating the combinations given access to randomly ordered explanations or the explanation chosen \emph{a priori} by the participant as their favourite. The differences in mean performance are statistically significant with a standard test rejecting a null hypothesis of equality with a \emph{p} value $<0.01$. The gap shows that Ardent allows for a more nuanced collaboration between human and AI such that the humans can really take advantage of a predictor that actually has a \emph{lower} accuracy on average than them, which may not be an obvious point when people evaluate the potential use of a decision support system.
The fact that Ardent outperforms random explanations provides evidence that a choice of explanations is important for people, and certainly validates that they can be very useful for giving them insight into a model's predictions.
In the end:
\emph{Ardent improves overall system performance by enabling useful human-AI collaboration.}

\textbf{Explanation Efficiency.}
By running posterior updates, Ardent incurs a computational cost, however this is not as large an issue as it may originally seem.
Given the more targeted explainer selection from Ardent, users actually click through $31.4\%$ fewer explanations on average, which saves on the computational cost of generating these explanations - which in some cases can require multiple passes through a network, potentially more than offsetting the cost of Ardent updates. Figure \ref{fig:viewed} shows more clearly how the average number of explanations viewed decreases over time with Ardent, increasing the efficiency. Interestingly, they also decrease for the Random group - given that there is no change in the way explanations are presented here, it appears that the main reason for this would be that the participants begin to fatigue of the task and are less inclined to click through explanations. 
It takes time to view, evaluate, and properly draw conclusions from an explanation and humans get less engaged as tasks go on, especially if they are repetitive. It is this aspect that Ardent aims to handle by producing a relative ordering. Ardent is then able to provide the most useful explanations first in order to engage the participant, but also is still able to offer alternative explanations when they are needed.
\emph{Targeted explanations can result in computational savings and decrease fatigue.}

\textbf{Preference Identification.}
In addition to the ability to optimise performance, Ardent obtains a ranking of which explainers users seem to find most useful - the ones that actually impact the behaviour of the human.
Figure \ref{fig:preferences} demonstrates the trajectory of an example user. It can be seen that in the beginning the selection of explainers is relatively random, as Ardent starts to learn which explainers are useful the ordering entropy decreases - Ardent identifies that this user finds the example-based methods most informative.
Importantly, Ardent outperforms the baseline arm that gives the participant the explanation that they \emph{a priori} thought would be the most useful. This emphasises how the impact of explainability is not as simple as a qualitative analysis of a method, and that what we think may be useful may not actually lead to significant change in the way that people come to decisions.
\emph{Ardent efficiently identifies individual preferences, potentially better than the individuals themselves.}

\vspace{-3mm}
\section{Discussion}
\vspace{-2mm}

In this work we introduced Ardent, an approach for optimal human-AI collaboration. Here we focus on high stakes settings where it is important for humans to remain in control while giving the support systems opportunities to convince them to pay attention when appropriate - this is validated through simulation as well as a study on image classification.
Ardent offers a solution when there is disagreement between the human and the decision support system, but does implicitly assume that at least one of them is correct. There are still many interesting directions that can be taken, especially building around a system like Ardent using semi/self-supervised learning to understand when/where both policies fail. There are many ways support systems can  empower human decision makers and we by no means expect Ardent to be the only component in a fully deployed meta-system.
Our hope is that Ardent will encourage and support the development of machine learning methods that work with people to provide the best of both worlds while remaining safe to deploy in challenging scenarios.

\section*{Acknowledgements}
AJC would like to acknowledge and thank Microsoft Research for its support through its PhD Scholarship Program with the EPSRC.
This work was additionally supported by the Office of Naval Research (ONR) and the NSF (Grant number: 1722516).


\bibliography{references}
\bibliographystyle{apalike}

\newpage
\appendix
\section{Experimental Setup}
\label{appdx:interaction}

\begin{figure*}[ht]
\centering
\includegraphics[width=1.0\textwidth]{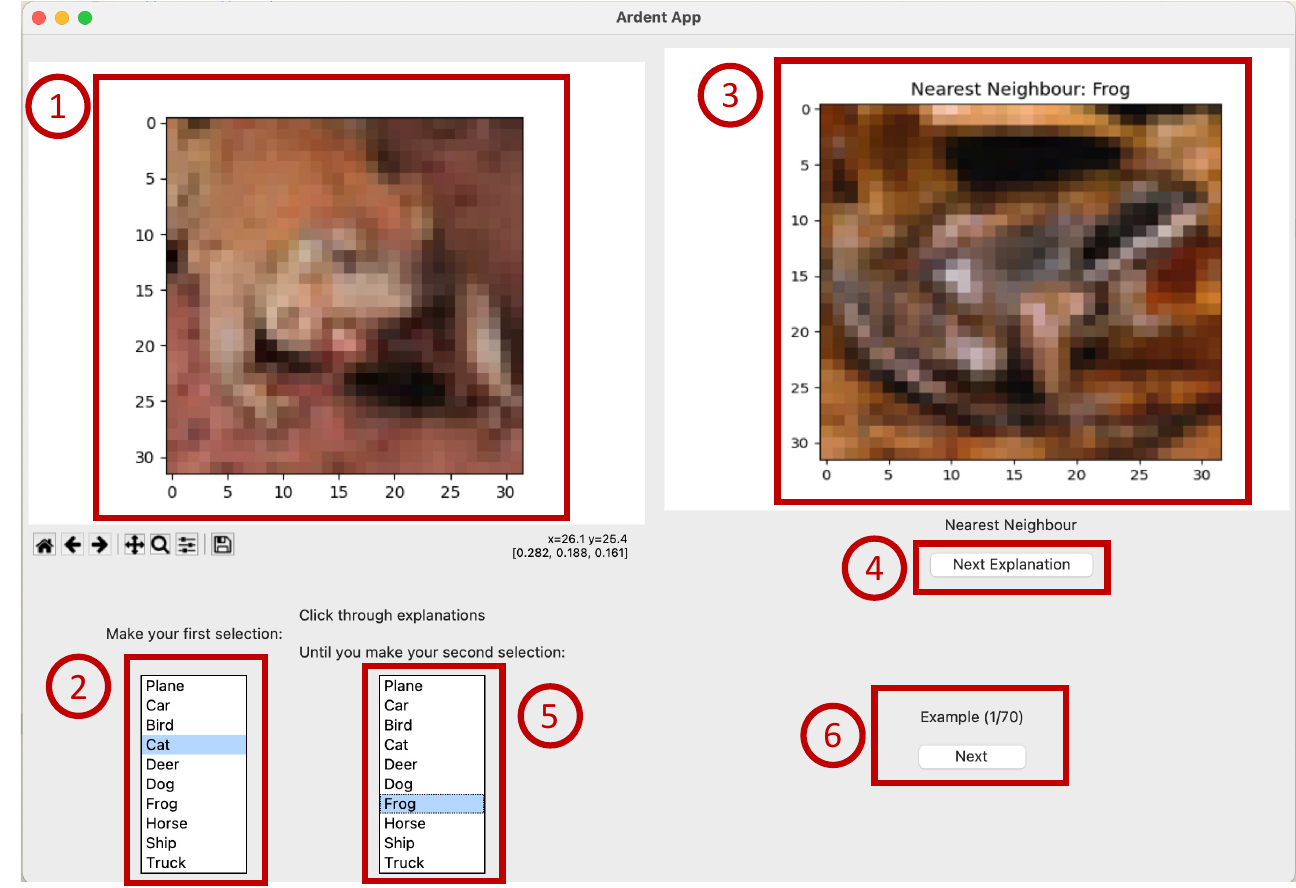}
\vspace{-4mm}
\caption{\textbf{Experimental presentation to participants.} Subjects are shown a new test image, and asked to make a prediction.
The system then shows them the model prediction, in this case `Frog', and as long as the subject remains unconvinced, continues to show them new explanations - here the user is being shown the nearest neighbour from the training set in latent space, which is a frog, and this has convinced the participant that the prediction might be right, despite previously thinking the image was of a ginger cat.}
\label{fig:gui} 
\vspace{-4mm} 
\end{figure*}

\subsection{Graphical User Interface}

The task is presented to the participants as in Figure \ref{fig:gui}, made up of the individual components that allow for interaction explained here:
\begin{enumerate}
    \item The test image that the participant is asked to classify.
    \item The participant is asked to select their first choice as to which is the correct classification. 3 would not be revealed at this point.
    \item Explanations appear in the top right as requested by the participant - here is shown an example of the nearest neighbour to the test example.
    \item While the participant remains unconvinced they can move to the next explanation by clicking this button.
    \item If and when the participant decides to change their answer they make a second selection here.
    \item The participant can end the interaction by pressing this button which takes them to the next example.
\end{enumerate}

The 70 test-set indices used for construction of the task were: \{  5,  15,  32,  33,  34,  46,  61,  65,  68,  74,  84,  86,  91, 100, 111, 115, 121, 126, 130, 134, 146, 163, 165, 169, 170, 183, 184, 187, 206, 223, 224, 228, 246, 248, 250, 254, 264, 266, 271, 275, 305, 309, 312, 313, 322, 323, 324, 340, 346, 356, 367, 385, 394, 418, 421, 426, 428, 439, 470, 481, 483, 493, 502, 511, 522, 531, 549, 572, 586, 610\}

\subsection{Participant Instructions}

Before completing the task, participants are shown the following information:

1. Introduction

You are invited to participate in a research study that aims to understand how machine learning methods affect human performance on image classification tasks. Before you decide to participate, it is important that you understand why the research is being conducted and what it will involve. Please take time to read the following information carefully.

2. Purpose of the Study

The purpose of this study is to investigate the effects of machine learning techniques on human performance in image classification tasks. We are interested in understanding how these methods can enhance or impact your ability to classify images accurately.

3. What Data Will Be Collected

During this study, we will collect data related to your performance in the image classification tasks, such as accuracy and response time. We will also gather basic demographic information such as age and gender. Please note that no sensitive data will be collected.

4. How the Data Will Be Used

The data collected will be used to assess the effectiveness of machine learning methods in enhancing human performance on the image classification task. The aggregated results may be published in academic journals, conference presentations, and technical reports. Individual responses will not be identifiable in any published or presented data.

5. How the Data Will Be Stored and for How Long

All data collected during the study will be securely stored in an encrypted format on secure servers. Data will be retained for a period of five years after the conclusion of the study, as required by our data retention policy, after which it will be securely deleted.

6. Anonymity of Responses

Your participation in this study will remain anonymous, using the randomised ID that has been assigned to you. No personally identifiable information will be associated with your responses in any reports of this research. The data will be presented in aggregate form.

7. Data Sharing with Other Researchers

Anonymised, aggregated data may be made available to other researchers online at some point. Again, individual responses will not be identifiable.

8. Withdrawal of Consent and Data

You have the right to withdraw from the study at any time. If you choose to withdraw, all data associated with your participation will be deleted. To withdraw your consent and data, please contact \textcolor{red}{[Redacted for double-blind review]} via email.

9. Legal Framework

Your data will be handled according to the principles and rules set by the General Data Protection Regulation (GDPR).

10. Consent

Please confirm that you have read and understand the above information relating to your participation in this research study.
By clicking the box below, you confirm that you:
\begin{itemize}
    \item Understand the nature and purpose of the study.
    \item Agree to the collection, use, and storage of your data as described above.
    \item Understand that your participation is voluntary and you may withdraw at any time without penalty.
    \item I agree to participate in this study
\end{itemize}

\section{Alternate Uses}
\textbf{Ardent for Education?}
By trying to find convincing explanations of the machine learning system, it could be thought that Ardent represents a method for education of the human expert. While a byproduct of the system may be that the human learns something when shown predictions and explanations in certain contexts,
it would be wrong to equate this to typical education methods \cite{luan2021review}. The setting in education is essentially to assume that $\pi_\mathrm{support}$ is the correct policy and thus try to minimise some divergence between the human and machine by \emph{influencing} them in some way \cite{korkmaz2019review}. This overlooks the case when the human is correct and the system is not, which as we establish is a very important aspect when it comes to the safety of any deployed system.
Ardent can be seen as taking the education-based approach to trying to determine the use of explainers. We determine if they were beneficial by measuring performance on the task - in the same way students are tested on their knowledge, not just asked the yes/no question of if they learnt something.

\end{document}